\RequirePackage[pagebackref=true,breaklinks=true,letterpaper=false,colorlinks,bookmarks=false]{hyperref} 
\documentclass[a4paper, 10pt, conference]{ieeeconf}      
\usepackage{FG2021}
\usepackage{amsmath,graphicx}
\usepackage{amssymb}
\usepackage{array}
\usepackage{fancybox}
\usepackage{fancyhdr}
\usepackage{float}
\usepackage{hhline}
\usepackage{longtable}
\usepackage{multirow}
\usepackage{comment}
\usepackage{tcolorbox}
\usepackage{xcolor}
\usepackage[utf8]{inputenc}

\FGfinalcopy 

\IEEEoverridecommandlockouts                              
\def\FGPaperID{****} 

\title{\LARGE \bf
Understanding Cross Domain Presentation Attack Detection for Visible Face Recognition
}

\author{\parbox{16cm}{\centering
    {\large Jennifer Hamblin\hspace{0.5in}Kshitij Nikhal\hspace{0.5in}Benjamin S. Riggan}\\
    {\normalsize
    Department Electrical and Computer Engineering, University of Nebraska-Lincoln, Lincoln, USA\\}}
    \thanks{}
}

\begin{document}
\IEEEoverridecommandlockouts\pubid{\makebox[\columnwidth]{978-1-6654-3176-7/21/\$31.00~\copyright{}2021 IEEE \hfill} \hspace{\columnsep}\makebox[\columnwidth]{ }}

\ifFGfinal
\thispagestyle{empty}
\pagestyle{empty}
\else
\author{Anonymous FG2021 submission\\ Paper ID \FGPaperID \\}
\pagestyle{plain}
\fi
\maketitle

\begin{abstract}


Face signatures, including size, shape, texture, skin tone, eye color, appearance, and scars/marks, are widely used as discriminative, biometric information for access control.  Despite recent advancements in facial recognition systems, presentation attacks on facial recognition systems have become increasingly sophisticated.  
The ability to detect presentation attacks or spoofing attempts is a pressing concern for the integrity, security, and trust of facial recognition systems.  Multi-spectral imaging has been previously introduced as a way to improve presentation attack detection by utilizing sensors that are sensitive to different regions of the electromagnetic spectrum (e.g., visible, near infrared, long-wave infrared).  Although multi-spectral presentation attack detection systems may be discriminative, the need for additional sensors and computational resources substantially increases complexity and costs.  Instead, we propose a method that exploits information from infrared imagery during training to increase the discriminability of visible-based presentation attack detection systems.  
We introduce (1) a new cross-domain presentation attack detection framework that increases the separability of bonafide and presentation attacks using only visible spectrum imagery, (2) an inverse domain regularization technique for added training stability when optimizing our cross-domain presentation attack detection framework, and (3) a dense domain adaptation subnetwork to transform representations between visible and non-visible domains.

\end{abstract}

\section{INTRODUCTION}

\label{sec:intro}

Facial recognition is a prevalent type of biometric authentication for consumer devices (e.g., smartphone, tablets, and laptop computers) and secured access points in public, private, and government facilities.  Attempts to bypass such securities by either 
spoofing (i.e., imitating) biometrics \cite{8553003} associated with an identity that has privileged access or prohibiting recognition capabilities (e.g., altering facial appearance) to avoid detection are considered presentation attacks (PAs).  Deep learning networks are nearing human level performance on face recognition tasks~\cite{GUO2019102805} in some respects, but vulnerability to PAs limits deployment in certain settings without some level of physical security~\cite{SOUZA2018368}. Therefore, presentation attack detection (PAD)---the ability to distinguish between bonafide (or genuine) faces and attacks---is critical for protecting private information (e.g., personal, financial, or proprietary information), increasing public safety, and furthering societal trust of facial recognition technology.  
Face PAs can include complete obfuscations, such as printed image, video replay, or mask style attacks, or partial obfuscations, such as wearing glasses, make-up, or wigs.  The partial obfuscations are generally more challenging to detect because these attacks are often acceptable societal behaviors and practices.  However, without extraneous non-visible cameras, such as depth, near infrared (NIR), or longwave infrared (thermal), the full obfuscation PAs are still rather difficult to detect since they are designed to accurately mimic the visual geometry and texture of actual faces.

Despite the benefits of combining information from multi-modal cameras for PAD applications, the added cost and complexity of using multiple sensors (e.g., visible, NIR, thermal, and depth) severely limits the use of PAD to local controlled access environments.  Therefore, to leverage current (and future) surveillance camera infrastructure that is mostly comprised of visible cameras, we aim to learn to estimate discriminative information (e.g., factors extracted from infrared imagery) from visible imagery using new domain adaptation objectives.

Recently, \cite{MCCNN} introduced the Wide Multi-Channel presentation Attack (WMCA) dataset that contains both 2D and 3D presentation attacks with spatially and temporally synchronized imagery across four different sensor domains.  The WMCA dataset contains eight different kinds of presentation attack instruments (PAIs) that fall under four main categories. These attack categories include facial disguise (plastic halloween masks, paper glasses, funny eye glasses), fake face (mannequin, flexible silicon masks, paper masks), photo (print/electronic images), and video.

Other multi-modal PAD datasets include: Casia-Surf~\cite{CasiaSurf}, MLFP~\cite{MLFP_2017_CVPR_Workshops}, Multispectral-Spoof (MSSpoof)~\cite{MSspoof}, 3DMAD~\cite{3DMAD}, as well as Casia-Surf CeFa~\cite{Liu_2021_WACV}. Many approaches utilize information from all available imaging modes in order to carry out the PAD task, which requires complex and expensive sensor suites to perform PAD.

Previous approaches (e.g., \cite{LBPcnn}) have used texture-based methods---local binary patterns (LBP)---to address 2D attacks, like print or video replay. LBP image representations have been primarily applied to visible imagery for PAD since infrared and depth imagery exhibit relatively fewer high frequency details (e.g., texture) compared with visible images.  Thus, LBP representations are suboptimal for domain adaptation between infrared and visible modalities.   


In this work, we enhance the performance of PAD systems that utilize readily available visible spectrum cameras and equipment by harnessing the auxiliary information present in supplementary image domains during the training process. However, NIR cameras with filters and thermal cameras need not be present at deployment.  Our primary contributions include:
\begin{itemize}
    \item Cross-Domain PAD (CD-PAD)---a task-level cross domain optimization process using thermal or NIR imagery to enhance discriminability of bonafide and PAs from visible spectrum imagery,
    \item Inverse Domain Regularization (IDR)---a new inverse domain regularization function,
    \item Dense Domain Adaptation (DDA) subnetwork---an improved method for learning the transformation from visible to the target domain,
    \item extensive analysis using the WMCA and MSSpoof datasets,
    \item an ablation study comparing our proposed IDR with other domain confusion methods.
\end{itemize}

Our proposed framework, including CD-PAD, IDR, and DDA, enhances visible-based PAD performance over \cite{MCCNN} by learning to predict information from discriminative infrared imagery from visible imagery during development.

    \begin{figure*}[htb]
      \centering
      \includegraphics[scale=.52]{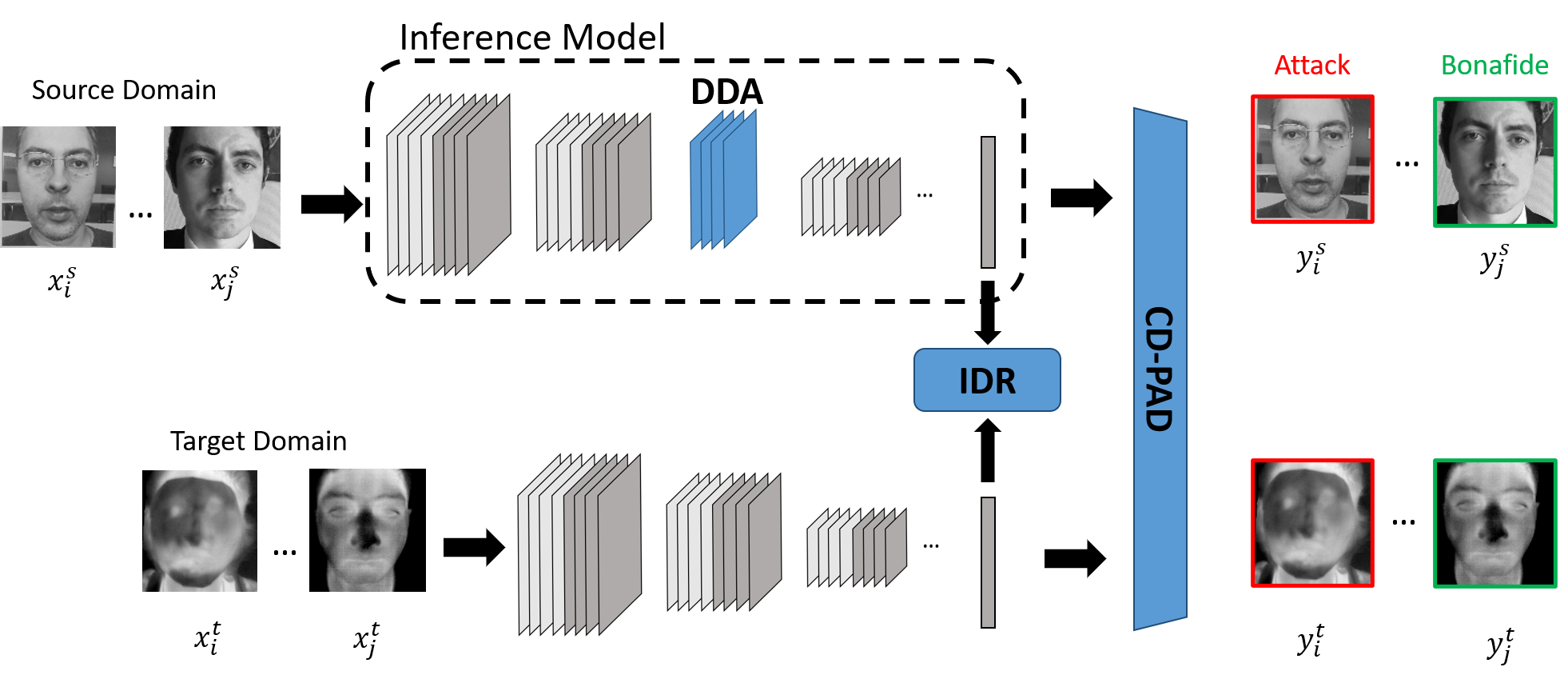}

      \caption{The proposed PAD framework exploits both Source and target domains to jointly optimize the base architecture and dense domain adaptation (DDA) subnetwork offline (i.e., during training) using the proposed cross-domain PAD (CD-PAD) objective function and inverse domain regularization (IDR).  However, only the inference model (source domain stream) is used during deployment since the goal is to limit operational complexity and cost of PAD systems.}
      \label{fig:diagram}
   \end{figure*}

\section{BACKGROUND}
\label{sec:background}

While there is a significant amount of research on liveness detection \cite{7900300,REHMAN2018159} and multi-modal PAD \cite{wmcaOneClass,Wang_2019_CVPR_Workshops,WMCAautoencoder}, there is limited work on using principles of domain adaptation for PAD.  Therefore, we summarize the Multi-Channel Presentation Attack Detection framework and domain adaptation techniques, including Maximum Mean Discrepancy (MMD), siamese networks, and domain invariance loss (DIL).

\subsection{Multi-Channel Presentation Attack Detection}
In \cite{wmcaOneClass,MCCNN} the Multi-Channel Convolutional Neural Network (MCCNN) was introduced for PAD.  First, \cite{MCCNN} proposed a multi-channel (i.e., multi-modal) fusion approach that combined information from four imaging modalities: visible, near infrared (NIR), longwave ``thermal'' infrared (LWIR), and depth  to perform PAD using the MCCNN architecture.  Then, in \cite{wmcaOneClass} the MCCNN is used to address the concern of novel ``unseen" attacks.  For the same purpose, \cite{WMCAautoencoder} developed an autoencoder network that utilizes the WMCA dataset to perform anomaly-based spoof detection.  The fundamental difference between these approaches and our work is that they exploit multi-modal imagery during inference.  Instead, we exploit multi-modal imagery offline in order to enhance the discriminability of visible-based PAD.

\subsection{Maximum Mean Discrepency}
\label{ssec:mmd}
The Maximum Mean Discrepancy (MMD)~\cite{MMD} is a measure that was proposed to evaluate the similarity between two distributions by computing distance between their reproducing kernel Hilbert space (RKHS) embeddings.  MMD has been used as a metric for minimizing the distance between source and target domain representations~\cite{Long_2013_ICCV,BeyondWS}. 3D CNN framework for PAD tasks was introduced in ~\cite{8335313} that incorporated MMD regularization between dataset domains for improved generalization.

Let $S=\{s_1,...,s_N\}$ and $T=\{t_1,...,t_N\}$ be the sets of features of the source and target domains.  In this particular problem each set has the same number of elements $N$, although in general that need not be the case.  Then the squared MMD of $S$ and $T$ can be expressed as 

\begin{equation}
\begin{aligned}
MMD^2(S,T) = &\sum_{i,j}^N \frac{k(s_i,s_j)}{N^2} + \sum_{i,j}^N \frac{k(t_i,t_j)}{N^2} \\&- 2\sum_{i,j}^N \frac{k(s_i,t_j)}{N^2}, 
\end{aligned}
\end{equation}
where $k(\cdot,\cdot)$ is the kernel associated with the RKHS.

The main disadvantage of MMD is that there is no discrimination between bonafide and attack instances.  Therefore, we investigate an alternative to MMD for domain adapation in the context of PAD for facial recognition systems.

\subsection{Siamese Networks for PAD}
\label{ssec:siamese}

Siamese networks~\cite{doi:10.1142/S0218001493000339} have been used to tackle both domain adaptation~\cite{Motiian_2017_ICCV}~\cite{8565895} and PAD tasks~\cite{Perez-Cabo_2019_CVPR_Workshops}.  In~\cite{8565895}, a siamese network implementing contrastive loss is used for heterogenous face recognition between different imaging domains where images are mapped to a shared embedding space.  Siamese networks work well when imaging domains are sufficiently close (see \cite{8100203}).  However, when imaging domains are further apart, they have been shown to under-perform.  Moreover, siamese networks, which are trained to ideally perform well on multiple domains at the same time, end up performing sub-optimally in all domains.  Instead, we focus on modeling the complex interrelationships between two domains for PAD.

\subsection{Domain Invariance Loss}
\label{ssec:domaininvariantloss}

The Domain Invariance Loss (DIL)~\cite{fondje2020crossdomain}\cite{Poster_2021_WACV} is a regularization technique proposed for domain adaptation for thermal-to-visible facial recognition tasks.  DIL uses a domain classification network that learns the probability that the features produced from an image belong to either the visible ($P_{vis}$) or thermal domain ($P_{therm}$).  Since the ultimate goal is increasing the similarity between the visible and thermal representations, the domain classifier is trained such that the two distributions are indistinguishable from each other.  Specifically, the domain classification labels are constant, i.e., $P_{vis}=P_{therm}=0.5$.  The potential disadvantage of this approach is that the labels are always the same, which implies there is a risk of models never learning patterns associated with either domain.  Therefore, we consider an alternative where such patterns are learned and used in a regularizing fashion.

\section{METHODOLOGY}
\label{sec:methodology}


The proposed PAD framework (Fig.~\ref{fig:diagram}) aims to enhance visible-based PAD using new domain adaptation principles.  First, we define the problem: cross-domain presentation attack detection (CD-PAD). Then, we introduce the components of our framework: (1) a base network architecture, (2) a new dense domain adaptation (DDA) subnetwork, (3) a new CD-PAD objective function, and (4) a new inverse domain regularization function.

\subsection{Preliminaries}
\label{ssec:prelim}
Enhancing PAD performance from visible spectrum imagery requires exploitation of subtle cues (e.g., specular reflections) to differentiate between bonafide faces and PAs. To emphasize such subtle cues, we introduce a new CD-PAD framework.
The CD-PAD problem is where discriminative information from a target domain is used to boost the quality of information extracted from the source domain. For example, by predicting infrared image representations from visible imagery, CD-PAD significantly improves the quality of visible-based PAD and reduces PAD system complexity (e.g. number/type of sensors) and cost.

Let $S=\{x^s_1, x^s_2, \dots, x^s_n\}$ and $T=\{x^t_1, x^t_2, \dots, x^t_m\}$ denote the sets of source (e.g., visible) and target (e.g., infrared) domain images, respectively. Here, $n$ is the number of images from the source domain and $m$ is the number of images from the target domain.

Let $(x^s_{i},x^t_{j})$ denote a pair of source and target images with corresponding labels $y^s_i$ and $y^t_j$.  Unlike methods that use restrictive Euclidean distance metrics to bridge domain gaps, CD-PAD performs inference level domain adaptation which relaxes the requirements for precise image registration/alignment and synchronous acquisition. Instead, the key requirement for CD-PAD is that both source and target labels sets, denoted by $\mathcal{Y}^s$ and $\mathcal{Y}^t$ respectively, must have overlapping labels.  Mathematically, this requirement is
\begin{equation}
    \mathcal{Y}^t \supseteq \mathcal{Y}^s,
    \label{eq:supset}
\end{equation}
where $y^t\in\mathcal{Y}^t$ and $y^s\in\mathcal{Y}^s$.

The main goal under our proposed CD-PAD framework is to learn a target domain PAD classifier, $P(y^t|f_t(x^t_j))$ that is sufficiently discriminative when used with source domain data, i.e., $P(y^t|f_s(x^s_i))$, where 
$f_{t}$ is the mapping from the target domain to the associated latent subspace and $f_{s}$ maps source imagery to the same ``target'' latent subspace. 
The primary objective for CD-PAD is to find an optimal source-to-target mapping $f_s$, such that $f_{s}(x^s_{i}) \approx f_{t}(x^t_{j})$.

\subsection{Base Architecture}

\begin{table}[htb]
\centering
\caption{Light CNN Architecture for $124 \times 118$ pixel image}

 \label{tab:LCNN}
\begin{tabular}{ |c|c|c|c|} 

 \hline
 Layer & Filter Size & Output Shape & Params \\ 
 & /Stride & ($H \times W \times C$)& \\\hline
    Conv1 & $5\times5/1$ & 124 $\times$ 118 $\times$ 96& 2,496\\
    MFM1 & --- & 124 $\times$ 118 $\times$ 48 & --\\ \hline
    Pool1 & $2\times2/2$ & 62 $\times$ 59 $\times$ 48 & -- \\ \hline
    Resblock1 & $\begin{bmatrix}
3\times3/1 \\
3\times3/1
\end{bmatrix}\times 1$ &  62 $\times$ 59 $\times$ 48 & 83,136 \\  
    \hline
    Conv2a & $1\times1/1$ & 62 $\times$ 59 $\times$ 96 & 4,704 \\
    MFM2a & --- &  62 $\times$ 59 $\times$ 48 & -- \\
    Conv2 & $3\times3/1$ &  62 $\times$ 59 $\times$ 192 & 83,136 \\
    MFM2 & --- &  62  $\times$59 $\times$ 96 & -- \\  \hline
    Pool2 &  $2\times2/2$ &  31 $\times$ 30 $\times$ 96 & -- \\ \hline
    Resblock2 &$\begin{bmatrix}
3\times3/1 \\
3\times3/1
\end{bmatrix}\times 2$    & 31$\times$ 30 $\times$ 96& 332,160 \\  
    \hline
    Conv3a & $1\times1/1$ & 31 $\times$ 30 $\times$ 192 & 18,624 \\
    MFM3a & --- & 31 $\times$ 30 $\times$ 96 & -- \\
    Conv3 & $3\times3/1$ & 31 $\times$ 30 $\times$ 384 & 332,160 \\
    MFM3 & --- & 31 $\times$ 30 $\times$ 192 & -- \\ \hline
    Pool3 & $2\times2/2$ & 16 $\times$ 15 $\times$ 192 & -- \\ \hline
    Resblock3 & $\begin{bmatrix}
3\times3/1 \\
3\times3/1
\end{bmatrix}\times 3$ & 16 $\times$ 15 $\times$ 192 & 1,327,872 \\  
    \hline
    Conv4a & $1\times1/1$ & 16 $\times$ 15 $\times$ 384  & 74,112\\
    MFM4a & --- & 16 $\times$ 15 $\times$ 192 & -- \\
    Conv4 & $3\times3/1$ & 16 $\times$ 15 $\times$ 256 & 442,624\\
    MFM4 & --- &  16 $\times$ 15 $\times$ 128  & --\\ \hline
    Resblock4 & $\begin{bmatrix}
3\times3/1 \\
3\times3/1
\end{bmatrix}\times 4$ & 16 $\times$ 15 $\times$ 128 & 590,336 \\ 
    \hline
    Conv5a & $1\times1/1$ & 16 $\times$ 15 $\times$ 256 & 33,024 \\
    MFM5a & --- & 16 $\times$ 15 $\times$ 128 & -- \\ \hline
    Conv5 & $3\times3/1$ & 16 $\times$ 15 $\times$ 256 & 295,168 \\
    MFM5 & --- &  16 $\times$ 15 $\times$ 128 & -- \\ \hline
    Pool4 & $2\times2/2$ & 8 $\times$ 8 $\times$ 128 & -- \\ \hline
    Linear & --- & 512 & 4,194,816 \\
    MFM6 & --- & 256 & -- \\
    \hline
    
\end{tabular}
\end{table}

The inclusion of additional spectral data has been shown to increase the discriminative power of multi-modal PAD systems~\cite{MCCNN}.  However, many extant security systems employ visible spectrum cameras and use visible enrollment face imagery.    Therefore, we propose a method that consists of training a PAD network to extract discriminative (e.g, infrared) features from visible imagery while only leveraging non-visible information during training.  The network contains two nearly identical data streams---one for source imagery and one for target imagery---consisting of CNNs with architectures based on the Light CNN network~\cite{LightCNN}.  Both streams are fed into the proposed CD-PAD classifier. The source stream, also referred to as the inference model in Fig.~\ref{fig:diagram}, is modified to include the addition of the DDA subnetwork described in section \ref{ssec:dda}.

The Light CNN is pre-trained on the MSCeleb-1M dataset \cite{LightCNN}, which is a large-scale face dataset.  Then, transfer learning is applied to both streams to re-use relevant model parameters.  This approach is similar to \cite{MCCNN}, except that our two stream network is trained in a domain adaptive manner instead of a multi-modal fusion manner.  Table~\ref{tab:LCNN} summarizes each layer of the Light CNN architecture, which is comprised of convolution (Conv), Max-Feature-Map (MFM), max pooling, and residual layers.

\subsection{DDA Sub-network}
\label{ssec:dda}

A domain adaptive subnetwork is added to the visible (source) stream of the CD-PAD network to learn the mapping from the source to target domain.  
We propose a new Dense Domain Adapation (DDA) subnetwork which is composed of a dense block \cite{DenseHuang_2017_CVPR} that consists of four convolutional layers as shown in Table \ref{tab:subnetarch}.  Mathematically, the DDA subnetwork is represented as
\begin{equation}
\begin{aligned}
    \delta(u) = Concat\{& \delta_{conv1}(u), \delta_{conv2}(u),\\ 
    & \delta_{conv3(u)}, \delta_{conv4}(u)\}, 
\end{aligned}
\label{eq:dda}
\end{equation}
where 
\begin{align}
    \delta_{conv1}(u) &= ReLU(Conv(BatchNorm(u))),\\
    \delta_{conv2}(u) &= ReLU(Conv(\delta_{conv1}(u))),\\
    \delta_{conv3}(u) &= ReLU(Conv(\delta_{conv2}(u))),\\
    \delta_{conv4}(u) &= ReLU(Conv(\delta_{conv3}(u))),
\end{align}
 with $Conv(\cdot)$ representing a $3\times3$ convolution and $ReLU(\cdot)$ the rectified linear unit activation function.
The parameters of the DDA subnetwork are optimized using our proposed CD-PAD loss (section~\ref{ssec:cdpad}).  

The DDA subnetwork (\ref{eq:dda}) is motivated by the Residual Spectrum Transform (RST) subnetwork in \cite{fondje2020crossdomain} that used a residual transformation \cite{REsnetHe_2016_CVPR} based subnetwork to bridge domain gaps for thermal-to-visible face recognition.  The effects of subnetwork type (i.e., residual versus dense) and placement within the Light CNN on overall performance of CD-PAD are described in Section \ref{ssec:ablationstudy}.  The dense architecture was selected for the DDA subnetwork primarily due to superior performance observed in the context of PAD.

\renewcommand{\tabcolsep}{1pt}
\begin{table}[htb]
\centering
\caption{DDA Subnetwork Architecture for $124 \times 118$ pixel image}

 \label{tab:subnetarch}
\begin{tabular}{ |c|c|c|c|c| } 

 \hline
 Layer & Inputs & Output Shape & Params \\ 
 \hline
  $BatchNorm$ & Pool3 & $16\times15\times192$ & 384\\
  \hline
  $\delta_{conv1}$ & $BatchNorm$ & $16\times15\times48$ & 82,992\\ 
  \hline
  $\delta_{conv2}$ & $\delta_{conv1}$ & $16\times15\times48$ & 20,784 \\ 
  \hline
  $\delta_{conv3}$ & [$\delta_{conv1}, \delta_{conv2}$] & $16\times15\times48$ & 41,520 \\ 
  \hline
  $\delta_{conv4}$ & [$\delta_{conv1}, \delta_{conv2}, \delta_{conv3}$] & $16\times15\times48$ & 62,256 \\ 
  \hline
  $\delta$ & [$\delta_{conv1}, \delta_{conv2}, \delta_{conv3}, \delta_{conv4}$] & $16\times15\times192$ & --- \\
 \hline
\end{tabular}
\end{table}
\renewcommand{\tabcolsep}{6pt}

The DDA subnetwork receives the output of the Pool3 max pooling layer shown in the Light CNN architecture in Table~\ref{tab:LCNN} as input to the BatchNorm layer.  The dense output of DDA is then passed to the Resblock3 layer of Light CNN and through the remainder of the network. 

\subsection{Cross Domain Presentation Attack Detection}
\label{ssec:cdpad}




Our proposed CD-PAD framework alternates training between domain modalities so that information extracted from the target domain face imagery can be learned and then used to guide the adaptation of source domain representations.  First, the PAD classifier is trained exclusively on the target data.  The target domain classifier and Light CNN are trained in a manner to avoid over-fitting to the target data.  We found that over-training on target data often leads to under-performing on source imagery (i.e., visible based PAD).   
The classifier weights are learned using the Binary Cross Entropy (BCE),  
\begin{equation}
\begin{aligned}
    \mathcal{L}(x^t,y^t)=&(1-y^t)\log(1-f(x^t; w^t))\\
    &+ y^t\log(f(x^t; w^t)).
\end{aligned}
\end{equation}

After this initial phase, the trained classifier is leveraged in order to adapt the DDA subnetwork that transforms the source imagery.  In the domain adaptive phase, the classifier weights, $w^t$, are fixed so that the objective function can only be minimized by transforming the feature representation of the visible domain.  For the domain adaptive training, the BCE loss function is
\begin{equation}
\begin{aligned}
    \mathcal{L}(x^s,y^s)=&(1-y^s)\log(1-f(x^s; w^t))\\
    &+ y^s\log(f(x^s; w^t)),
\end{aligned}
\end{equation}
where $w^t$ represents the classifier parameter weights that had previously been trained on the target data.

The CD-PAD framework ultimately works due to the fundamental assumption in (\ref{eq:supset}), where we assume that both target and source domain span the same label sets.  Due to the asynchronous, alternating training strategy used by CD-PAD, target and source imagery are not required to be precisely synchronized or co-registered.  Therefore, CD-PAD is more flexible and extensible than existing domain adaptation methodologies, especially those that optimize Euclidean distances between corresponding pairs or triplets. 


After training the CD-PAD framework, only the source stream (i.e., inference model in Fig.~\ref{fig:diagram}) is used for deployment of the PAD system.  This provides a very efficient and cost effective solution for PAD.  

\begin{table*}[htb]
\centering
 \caption{Subnetwork ablation study}
 \label{tab:subnet}
 \begin{tabular}{p{1.2cm}>{\centering}p{1.4cm}>{\centering}p{1.8cm}>{\centering}p{1.8cm}>{\centering}p{1.8cm}>{\centering\arraybackslash}p{1.8cm}}
 \hline
 \multicolumn{2}{c}{Network Details } & \multicolumn{2}{c}{Visible / Thermal} & \multicolumn{2}{c}{Visible / NIR}\\
 \hline
  Subnet Type & Layer & BPCER @1\% APCER & BPCER @5\% APCER& BPCER @1\% APCER & BPCER @5\% APCER\\ [0.5ex] 
 \hline
  None & No DDA  & 41.69 $\pm$ 17.32  & 34.67 $\pm$ 15.99 & 64.32 $\pm$ 2.79 & 50.92 $\pm$ 4.71   \\
  \hline
  \multirow{3}{*}{Dense} & Pool2  &  66.75 $\pm$ 8.45 & 53.32 $\pm$ 11.79 &   62.68 $\pm$ 9.99 & 71.47 $\pm$ 23.55 \\
   & Pool3 &  \textbf{29.64} $\pm$17.89 & \textbf{13.78} $\pm$ 7.11  &  \textbf{18.7} $\pm$ 1.77 & \textbf{9.95} $\pm$ 1.31  \\
   & Pool4  &  66.61 $\pm$ 12.17 & 51.97 $\pm$ 19.99  & 71.47 $\pm$ 23.55 & 55.66 $\pm$ 19.44  \\
 \hline
  \multirow{3}{*}{Residual} & Pool2 & 68.45 $\pm$ 10.99 & 45.11 $\pm$ 14.49 &  68.24 $\pm$ 12.9 & 50.11 $\pm$ 14.49 \\
   & Pool3 &  39.75 $\pm$ 14.09 & 18.99$\pm$ 10.17  & 20.69 $\pm$ 3.22 & 10.37 $\pm$ 1.06 \\
   & Pool4  &  86.92 $\pm$ 10.77 &  66.21$\pm$ 10.11  &  95.9 $\pm$ 1.05 & 79.64 $\pm$ 5.28  \\
   \hline
 \end{tabular}
 
\end{table*}

\subsection{Inverse Domain Regularization}
Lastly, we also propose a variation on the domain invariance loss~\cite{fondje2020crossdomain}.  In this variation, which we call inverse domain regularization (IDR), a domain classification network is instead trained to correctly differentiate between the imaging domains.  Inverting the labels of the source data is what drives the domain adaptation provided by IDR.  

First, the IDR domain classifier is trained with correct domain labels for each of the input images and learns to appropriately discriminate between the two domains. Using the same notation from Section \ref{ssec:prelim}, let $P^t(x_i)$ be the probability that a given training image $x_i$ comes from the target domain.  The IDR classifier is trained to predict $P^t(x^s_i)=0$ and $P^t(x^t_i)=1$, a correct classification of the domains for the respective inputs. Next, to guide the network to map the source images to the target domain, we implement the domain inversion of IDR.  
In this domain adaptive stage of training, the domain classifier parameters are fixed while the DDA subnetwork is updated.  
Here, the DDA network is also trained using using the IDR classifier, except the domain labels are inverted so that the source domain representations will appear to be similiar to the target domain representations. Mathematically, $P^t(x^s_i)=1.$
The bottom line is that IDR aims to reduce differences between source and target image representations in a class agnostic manner and thus complements the CD-PAD loss by imposing additional constraints.

\section{EXPERIMENTS}
\label{sec:experiments}
\subsection{Protocols}
\label{ssec:protocols}

\subsubsection{WMCA}
For training and evaluation on the WMCA dataset, the ``grandtest" protocol referred to in ~\cite{MCCNN} is used.  The data is split into three subsets: train, dev, and test.  For each domain, the subsets contain 28,223, 27,850, and 27,740 images respectively.  The distribution of attack categories are consistent across each of the sets and individual subjects do not appear in multiple subsets.  
The test subset contains 5,750 bonafide images, 1,649 facial disguise images, 13,041 fake face images, 4,200 photo attack images, and 3,100 video attack images.

\subsubsection{MSSpoof}
To show CD-PAD's potential for generalization, we also evaluate on the MSSpoof~\cite{MSspoof} dataset.  MSSpoof contains both visible and NIR imagery of 21 individuals.  Like WMCA, MSSpoof is split into three identity disjoint subsets: train, dev, and test.  All of the PAs in the MSSpoof dataset are print style attacks. The training subset contains 594 visible images and 577 NIR images, the dev subset contains 398 visible images and 395 NIR images, and the test subset contains 396 visible images and 395 NIR images.


\subsection{Implementation}
\label{ssec:implementation}

All models were trained in PyTorch~\cite{NEURIPS2019_9015} and updated using the ADAM optimizer~\cite{Kingma2015AdamAM} with a learning rate of 1x$10^{-4}$.  Features were generated using the Light CNN~\cite{LightCNN} framework initialized with weights pre-trained for facial recognition.  During the first phase of training, the fully-connected PAD classification layers are trained on thermal data. In the final cross domain training stage the weights of the DDA subnetwork in the visible data stream are made trainable.  The second stage of training uses the same optimizer and learning rate.  Networks trained with inverse domain regularization required an additional stage of training, wherein only layers in the parallel domain classification network are updated. Data augmentation is utilized during training with random horizontal flipping with a probability of 0.5, and random rotation of maximum 10 degrees.

Preprocessing on the MSSpoof dataset included 5-point facial landmark registration and tight cropping around the face.  Image cropping is utilized to alleviate potential problems with over-fitting as a result of the limited quantity of data in MSSpoof.  Restricting the network to only learn from information contained in the face prevents it from focusing on background details that are often dataset specific.

For reproducibility, our code is available at [removed for anonymity during review].

 \begin{figure*}[htb]
      \centering
      \includegraphics[scale=.65]{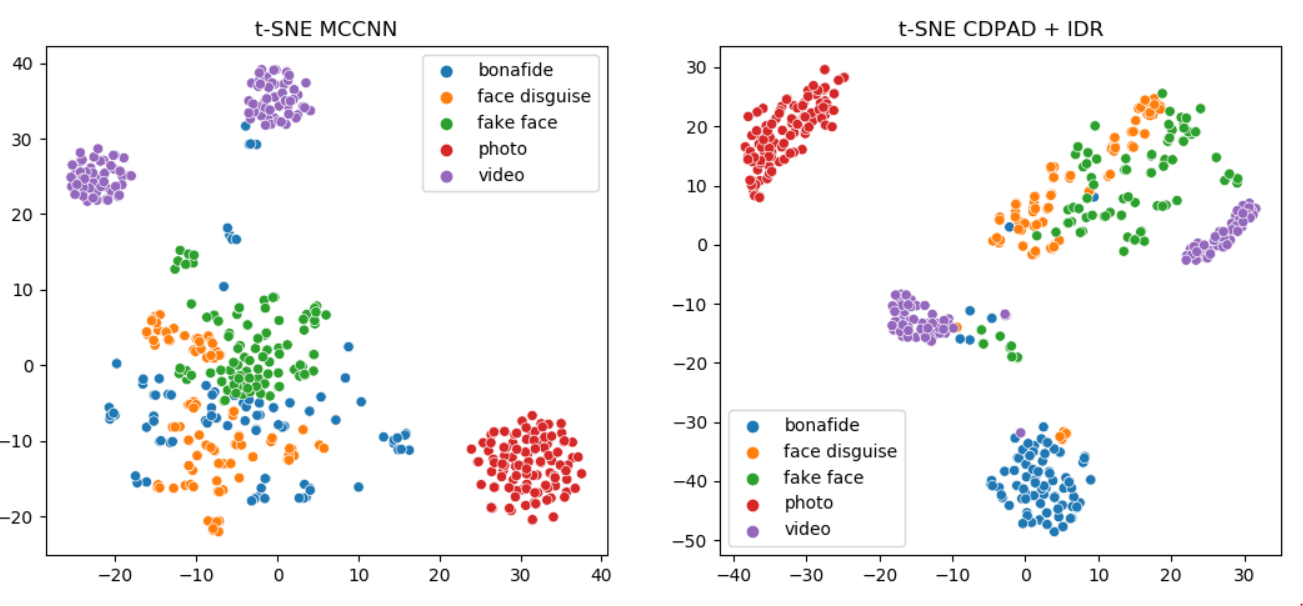}

      \caption{ Compared to the single domain visible baseline, our method shows better separability between bonafide and all attack data points.}
      \label{fig:tsne}
   \end{figure*}

\subsection{Evaluation Metrics}
\label{ssec:evaluation}
Results are reported according to the ISO/IEC 30107-3 standard metrics for presentation attack detection, Attack Presentation Classification Error Rate (APCER), Bonafide Presentation Classification Error Rate (BPCER), and the Average Classification Error Rate (ACER)~\cite{ISO}.  APCER designates the proportion of presentation attacks incorrectly identified as bonafide presentations, and BPCER is the proportion of incorrectly identified bonafide presentations.  The metrics are defined as follows:

\begin{equation} APCER(\tau) = \frac{FP(\tau)}{FP(\tau) + TN(\tau)}, 
\end{equation}

\begin{equation} BPCER(\tau) = \frac{FN(\tau)}{FN(\tau) + TP(\tau)}, \end{equation}
\begin{equation} ACER(\tau) = \frac{APCER(\tau) + BPCER(\tau)}{2}, \end{equation}
where FN, FP, TN, and TP are the number of false negatives, false positives, true negatives, and true positives for a given threshold $\tau$.

\subsection{Subnetwork Ablation Study}
\label{ssec:ablationstudy}



 

A subnetwork ablation study was conducted in order to determine the optimal layer depth at which to insert a domain adaptive subnetwork into the visible channel of the CD-PAD network.  The Light CNN network contains four max pooling layers that conclude each convolutional block.  In each test, a trainable subnetwork is placed directly after one of the max pooling layers to learn the transformation from the source to target domain.  When the subnetworks are used for domain adaptation, all pre-trained layers in the network remain fixed during training.

Table \ref{tab:subnet} shows the effects of subnetwork type (residual or dense) and location when using the CD-PAD method.  For this ablation study, additional regularizing loss functions are not implemented in the domain adaptive phase of training in order to highlight the change in performance that can be attributed to the network architecture alone.

In both cases, the domain adaptive subnetwork shows the greatest effect when placed after the third pooling layer in the Light CNN.  The most drastic improvements are seen in the lowest false positive rates where the CD-PAD network struggles without additional domain regularization.  All the final results in this paper are generated with the dense domain adaptive block at the third max pooling layer, which we refer to as the DDA subnetwork.



\subsection{Qualitative Analysis}
\label{ssec:qual}

To illustrate the enhancements due to CD-PAD, we evaluate the feature representations of the visible imagery (bonafide and PA samples) using the t-Distributed Stochastic Neighbor Embedding (t-SNE) \cite{tsne} method to visualize the data.  The t-SNE representations for the visible baseline and the CD-PAD adapted visible features are shown in Fig.~\ref{fig:tsne}.  Data samples used for the t-SNE visualization are randomly selected from the test set.  The adapted features are generated from a CD-PAD + IDR network trained with thermal imagery as the target domain.  It is clear from the plots that the cross domain adaptation causes the bonafide samples to be more tightly clustered in the feature space and have less overlap with the attack samples.

\subsection{Quantitative Results}
\label{ssec:results}

Next, we compare the performance of CD-PAD with visible and infrared (thermal or NIR) baseline models using WMCA and MSSpoof.  
The thermal and NIR specific models represent the upper performance bounds that can be attained by adapting visible data via our CD-PAD framework.  

\begin{table}[htb]
\centering
 \caption{CD-PAD results where NIR is the target domain using the WMCA dataset}
 \label{table:ir}
 \begin{tabular}{p{2cm}>{\centering}p{1.6cm}>{\centering}p{1.6cm}>{\centering\arraybackslash}p{1.7cm}}
 \hline
 Method  & BPCER @ 1\% APCER & BPCER @ 5\% APCER & AUC\\ [0.5ex] 
 \hline\hline

        MCCNN\cite{MCCNN}(NIR) & 5.93 $\pm$ 6.54 & 1.54 $\pm$ 1.94 & 0.997 $\pm$ 0.003 \\ \hline
        MCCNN\cite{MCCNN} (Visible) & 74.59 $\pm$ 9.87 & 43.72 $\pm$ 9.43 & 0.895 $\pm$ 0.029 \\ \hline
        Siamese network & 26.08 $\pm$ 3.16 & 11.18 $\pm$ 1.97 & 0.957 $\pm 0.008$ \\ \hline
        CD-PAD &  18.7 $\pm$ 1.77 & 9.95 $\pm$ 1.31  & 0.962 $\pm$ 0.008  \\ \hline
        CD-PAD+DIL & 19.84 $\pm$ 0.34 & 11.2 $\pm$ 1.41 & 0.970 $\pm$ 0.002 \\ \hline
        CD-PAD+MMD &  20.9 $\pm$ 4.1 & 13.1 $\pm$ 2.24 & 0.977 $\pm$ 0.001\\ \hline
        CD-PAD+IDR  &\textbf{ 17.13}$\pm$ 1.38 & \textbf{9.27} $\pm$ 2.13 & \textbf{0.980} $\pm$ 0.000 \\ \hline
    \hline
 \end{tabular}
\end{table}

\subsubsection{WMCA} We compare the results of the CD-PAD method using two different target domains, thermal and NIR, against networks trained for the PAD task on single modal data.   The CD-PAD method improves upon the quality of the attack detector when only visible data is available in a deployment scenario.  

The effects of using NIR imagery as the visible adaptation target are shown in Table \ref{table:ir}.  The CD-PAD network greatly improves over the visible baseline.  With NIR as the target domain CD-PAD achieves an average of 18.7\% BPCER at a 1\% APCER operating point, improving over the visible baseline by 55.89\%.  Adding IDR to the CD-PAD framework results in an additional improvement of 1.57\%.

Table \ref{table:therm} shows the results when thermal imagery is available for cross domain training. When CD-PAD is used on its own, the visible based PAD results are boosted.  CD-PAD shows a marked improvement in the BPCER at low APCER operating points in the ROC curve. CD-PAD achieves an average of 24.3\%  BPCER at a 1\% APCER operating point, and improves by 50.29\% over the visible baseline.  Including additional domain adaptation loss components had varying effects on the CD-PAD performance.  Introducing MMD to help guide domain adaptation actually hurt performance.  However, the combination of CD-PAD and IDR using the NIR target imagery achieved the biggest improvement in visible based PAD on WMCA decreasing the BPCER at a 1\% APCER by 62.17\%.  

\subsubsection{MSSpoof} In Table \ref{table:msspoof}, we evaluate the CD-PAD method using the MSSpoof dataset where visible source imagery is adapted to the target NIR domain.  Once again, CD-PAD improves upon training the model on visible imagery alone. The network trained on visible imagery achieves an average of 42.85\% BPCER at a 1\% APCER operating point, and the CD-PAD network achieves 13.75\% for the same metric.  Adding IDR to the CD-PAD framework offers a small performance boost to the BPCER score at 1\% APCER, improving the CD-PAD performance by 0.27\%.

\begin{table}[t]
\centering
 \caption{CD-PAD results where thermal is the target domain using the WMCA dataset}
 \label{table:therm}
 \begin{tabular}{p{2cm}>{\centering}p{1.6cm}>{\centering}p{1.6cm}>{\centering\arraybackslash}p{1.7cm}}
 \hline
 Method  & BPCER @ 1\% APCER & BPCER @ 5\% APCER & AUC\\ [0.5ex] 
 \hline\hline
        MCCNN\cite{MCCNN}(Thermal) &  3.83 $\pm$ 2.45 & 0.0 $\pm$ 0.0 & 0.998 $\pm$ 0.001 \\ \hline
        MCCNN\cite{MCCNN}(Visible) &  74.59 $\pm$ 9.87 & 43.72 $\pm$ 9.43 & 0.895 $\pm$ 0.029 \\ \hline
        Siamese network & 36.18 $\pm$ 3.17 & 19.89 $\pm$ 6.86 & 0.939 $\pm$ 0.014 \\ \hline
        CD-PAD & 24.3 $\pm$ 2.36 & 8.75 $\pm$ 1.64 & 0.973 $\pm$ 0.003 \\ \hline
        CD-PAD+DIL &  19.31 $\pm$ 1.25 & 7.88 $\pm$ 2.83 & 0.981 $\pm$ 0.006 \\ \hline
        CD-PAD+MMD & 48.24 $\pm$ 0.79 & 23.4 $\pm$ 1.98 & 0.948 $\pm$ 0.001 \\ \hline
        CD-PAD+IDR & \textbf{12.42} $\pm$  0.52 & \textbf{6.90} $\pm$ 1.56 & \textbf{0.982} $\pm$ 0.007 \\ \hline
    \hline
 \end{tabular}
 \end{table}

\begin{table}[htb]
\centering
 \caption{CD-PAD results for MSSpoof. NIR is the target domain.}
 \label{table:msspoof}
 \begin{tabular}{p{2cm}>{\centering}p{1.6cm}>{\centering}p{1.6cm}>{\centering\arraybackslash}p{1.7cm}}
 \hline
 Method  & BPCER @ 1\% APCER & BPCER @ 5\% APCER & AUC\\ [0.5ex] 
 \hline\hline
        MCCNN\cite{MCCNN}(IR) &  16.27 $\pm$ 3.74  & 12.09 $\pm$ 3.1 & 0.977 $\pm$ 0.003  \\ \hline
        MCCNN\cite{MCCNN}(Visible) & 42.85 $\pm$ 0.54 & 27.67 $\pm$ 0.51 & 0.891 $\pm$ 0.025 \\ \hline
        CD-PAD & 13.75 $\pm$ 0.59 & 9.25 $\pm$ 0.3 & 0.987 $\pm$ 0.001 \\ \hline
        CD-PAD+DIL &  14.74 $\pm$ 2.39 & \textbf{8.11} $\pm$ 0.81 & 0.987  $\pm$ 0.001 \\ \hline
        CD-PAD+MMD & 28.36 $\pm$ 3.76 & 17.54 $\pm$ 0.26 & 0.976 $\pm$ 0.037 \\ \hline
        CD-PAD+IDR & \textbf{13.48} $\pm$ 2.02 & 10.46 $\pm$ 2.64 & \textbf{0.987} $\pm$ 0.002 \\ \hline
    \hline
 \end{tabular}
 \end{table}


.  


\section{Conclusions}
In this paper, we proposed a new domain adaptation framework called CD-PAD that utilized mutli-modal data during training to improve visible based PAD for face recognition systems. 
To this end, we introduced (1) a new CD-PAD framework that increases the separability of bonafide and presentation attacks using only visible spectrum imagery, (2) an IDR technique for enhanced PAD and stability during optimization, and (3) a DDA subnetwork to transform representations between visible and infrared domains.  We found that our CD-PAD framework was able to significantly reduce the BPCER @ 1\% APCER by 57.46\%, 62.17\% and 29.37\% on the WMCA (NIR), WMCA (thermal), and MSSpoof (NIR) protocols.  Moreover, we found that our proposed IDR resulted in better PAD performance than previous MMD and DIL techniques.  The results imply that the CD-PAD framework is capable of providing very discriminative PAD while reducing the number/type of operation sensors, which enables less complex and more cost efficient PAD systems.




\addtolength{\textheight}{-3cm}   

{\small
\bibliographystyle{ieee}
\bibliography{egbib}

\begin{thebibliography}{10}\itemsep=-1pt

\bibitem{MLFP_2017_CVPR_Workshops}
A.~Agarwal, D.~Yadav, N.~Kohli, R.~Singh, M.~Vatsa, and A.~Noore.
\newblock Face presentation attack with latex masks in multispectral videos.
\newblock In {\em Proceedings of the IEEE Conference on Computer Vision and
  Pattern Recognition (CVPR) Workshops}, July 2017.

\bibitem{doi:10.1142/S0218001493000339}
J.~BROMLEY, J.~W. BENTZ, L.~BOTTOU, I.~GUYON, Y.~LECUN, C.~MOORE,
  E.~SÄCKINGER, and R.~SHAH.
\newblock Signature verification using a “siamese” time delay neural
  network.
\newblock {\em International Journal of Pattern Recognition and Artificial
  Intelligence}, 07(04):669--688, 1993.

\bibitem{MSspoof}
I.~Chingovska, N.~Erdogmus, A.~Anjos, and S.~Marcel.
\newblock {\em Face Recognition Systems Under Spoofing Attacks}, pages
  165--194.
\newblock Springer International Publishing, 2016.

\bibitem{8565895}
T.~de~Freitas~Pereira, A.~Anjos, and S.~Marcel.
\newblock Heterogeneous face recognition using domain specific units.
\newblock {\em IEEE Transactions on Information Forensics and Security},
  14(7):1803--1816, 2019.

\bibitem{LBPcnn}
G.~B. {de Souza}, D.~F. {da Silva Santos}, R.~G. {Pires}, A.~N. {Marana}, and
  J.~P. {Papa}.
\newblock Deep texture features for robust face spoofing detection.
\newblock {\em IEEE Transactions on Circuits and Systems II: Express Briefs},
  64(12):1397--1401, 2017.

\bibitem{3DMAD}
N.~{Erdogmus} and S.~{Marcel}.
\newblock Spoofing 2d face recognition systems with 3d masks.
\newblock In {\em 2013 International Conference of the BIOSIG Special Interest
  Group (BIOSIG)}, pages 1--8, 2013.

\bibitem{fondje2020crossdomain}
C.~N. Fondje, S.~Hu, N.~J. Short, and B.~S. Riggan.
\newblock Cross-domain identification for thermal-to-visible face recognition,
  2020.

\bibitem{wmcaOneClass}
A.~{George} and S.~{Marcel}.
\newblock Learning one class representations for face presentation attack
  detection using multi-channel convolutional neural networks.
\newblock {\em IEEE Transactions on Information Forensics and Security},
  16:361--375, 2021.

\bibitem{MCCNN}
A.~{George}, Z.~{Mostaani}, D.~{Geissenbuhler}, O.~{Nikisins}, A.~{Anjos}, and
  S.~{Marcel}.
\newblock Biometric face presentation attack detection with multi-channel
  convolutional neural network.
\newblock {\em IEEE Transactions on Information Forensics and Security},
  15:42--55, 2020.

\bibitem{MMD}
A.~Gretton, K.~Borgwardt, M.~Rasch, B.~Sch\"{o}lkopf, and A.~J. Smola.
\newblock A kernel method for the two-sample-problem.
\newblock In B.~Sch\"{o}lkopf, J.~C. Platt, and T.~Hoffman, editors, {\em
  Advances in Neural Information Processing Systems 19}, pages 513--520. MIT
  Press, 2007.

\bibitem{GUO2019102805}
G.~Guo and N.~Zhang.
\newblock A survey on deep learning based face recognition.
\newblock {\em Computer Vision and Image Understanding}, 189:102805, 2019.

\bibitem{REsnetHe_2016_CVPR}
K.~He, X.~Zhang, S.~Ren, and J.~Sun.
\newblock Deep residual learning for image recognition.
\newblock In {\em Proceedings of the IEEE Conference on Computer Vision and
  Pattern Recognition (CVPR)}, June 2016.

\bibitem{DenseHuang_2017_CVPR}
G.~Huang, Z.~Liu, L.~van~der Maaten, and K.~Q. Weinberger.
\newblock Densely connected convolutional networks.
\newblock In {\em Proceedings of the IEEE Conference on Computer Vision and
  Pattern Recognition (CVPR)}, July 2017.

\bibitem{ISO}
{Information technology — Biometric presentation attack detection}.
\newblock Standard, International Organization for Standardization, Geneva, CH,
  2017.

\bibitem{Kingma2015AdamAM}
D.~P. Kingma and J.~Ba.
\newblock Adam: A method for stochastic optimization.
\newblock {\em CoRR}, abs/1412.6980, 2015.

\bibitem{8100203}
J.~Lezama, Q.~Qiu, and G.~Sapiro.
\newblock Not afraid of the dark: Nir-vis face recognition via cross-spectral
  hallucination and low-rank embedding.
\newblock In {\em 2017 IEEE Conference on Computer Vision and Pattern
  Recognition (CVPR)}, pages 6807--6816, 2017.

\bibitem{8335313}
H.~{Li}, P.~{He}, S.~{Wang}, A.~{Rocha}, X.~{Jiang}, and A.~C. {Kot}.
\newblock Learning generalized deep feature representation for face
  anti-spoofing.
\newblock {\em IEEE Transactions on Information Forensics and Security},
  13(10):2639--2652, 2018.

\bibitem{7900300}
X.~Li, J.~Komulainen, G.~Zhao, P.-C. Yuen, and M.~Pietikäinen.
\newblock Generalized face anti-spoofing by detecting pulse from face videos.
\newblock In {\em 2016 23rd International Conference on Pattern Recognition
  (ICPR)}, pages 4244--4249, 2016.

\bibitem{Liu_2021_WACV}
A.~Liu, Z.~Tan, J.~Wan, S.~Escalera, G.~Guo, and S.~Z. Li.
\newblock Casia-surf cefa: A benchmark for multi-modal cross-ethnicity face
  anti-spoofing.
\newblock In {\em Proceedings of the IEEE/CVF Winter Conference on Applications
  of Computer Vision (WACV)}, pages 1179--1187, January 2021.

\bibitem{Long_2013_ICCV}
M.~Long, J.~Wang, G.~Ding, J.~Sun, and P.~S. Yu.
\newblock Transfer feature learning with joint distribution adaptation.
\newblock In {\em Proceedings of the IEEE International Conference on Computer
  Vision (ICCV)}, December 2013.

\bibitem{Motiian_2017_ICCV}
S.~Motiian, M.~Piccirilli, D.~A. Adjeroh, and G.~Doretto.
\newblock Unified deep supervised domain adaptation and generalization.
\newblock In {\em Proceedings of the IEEE International Conference on Computer
  Vision (ICCV)}, Oct 2017.

\bibitem{NEURIPS2019_9015}
A.~Paszke, S.~Gross, F.~Massa, A.~Lerer, J.~Bradbury, G.~Chanan, T.~Killeen,
  Z.~Lin, N.~Gimelshein, L.~Antiga, A.~Desmaison, A.~Kopf, E.~Yang, Z.~DeVito,
  M.~Raison, A.~Tejani, S.~Chilamkurthy, B.~Steiner, L.~Fang, J.~Bai, and
  S.~Chintala.
\newblock Pytorch: An imperative style, high-performance deep learning library.
\newblock In H.~Wallach, H.~Larochelle, A.~Beygelzimer, F.~d\textquotesingle
  Alch\'{e}-Buc, E.~Fox, and R.~Garnett, editors, {\em Advances in Neural
  Information Processing Systems 32}, pages 8024--8035. Curran Associates,
  Inc., 2019.

\bibitem{Perez-Cabo_2019_CVPR_Workshops}
D.~Perez-Cabo, D.~Jimenez-Cabello, A.~Costa-Pazo, and R.~J. Lopez-Sastre.
\newblock Deep anomaly detection for generalized face anti-spoofing.
\newblock In {\em Proceedings of the IEEE/CVF Conference on Computer Vision and
  Pattern Recognition (CVPR) Workshops}, June 2019.

\bibitem{8553003}
D.~Poster, N.~Nasrabadi, and B.~Riggan.
\newblock Deep sparse feature selection and fusion for textured contact lens
  detection.
\newblock In {\em 2018 International Conference of the Biometrics Special
  Interest Group (BIOSIG)}, pages 1--5, 2018.

\bibitem{Poster_2021_WACV}
D.~Poster, M.~Thielke, R.~Nguyen, S.~Rajaraman, X.~Di, C.~N. Fondje, V.~M.
  Patel, N.~J. Short, B.~S. Riggan, N.~M. Nasrabadi, and S.~Hu.
\newblock A large-scale, time-synchronized visible and thermal face dataset.
\newblock In {\em Proceedings of the IEEE/CVF Winter Conference on Applications
  of Computer Vision (WACV)}, pages 1559--1568, January 2021.

\bibitem{REHMAN2018159}
Y.~A.~U. Rehman, L.~M. Po, and M.~Liu.
\newblock Livenet: Improving features generalization for face liveness
  detection using convolution neural networks.
\newblock {\em Expert Systems with Applications}, 108:159--169, 2018.

\bibitem{BeyondWS}
A.~{Rozantsev}, M.~{Salzmann}, and P.~{Fua}.
\newblock Beyond sharing weights for deep domain adaptation.
\newblock {\em IEEE Transactions on Pattern Analysis and Machine Intelligence},
  41(4):801--814, 2019.

\bibitem{SOUZA2018368}
L.~Souza, L.~Oliveira, M.~Pamplona, and J.~Papa.
\newblock How far did we get in face spoofing detection?
\newblock {\em Engineering Applications of Artificial Intelligence}, 72:368 --
  381, 2018.

\bibitem{tsne}
L.~van~der Maaten and G.~Hinton.
\newblock Viualizing data using t-sne.
\newblock {\em Journal of Machine Learning Research}, 9:2579--2605, 11 2008.

\bibitem{Wang_2019_CVPR_Workshops}
G.~Wang, C.~Lan, H.~Han, S.~Shan, and X.~Chen.
\newblock Multi-modal face presentation attack detection via spatial and
  channel attentions.
\newblock In {\em Proceedings of the IEEE/CVF Conference on Computer Vision and
  Pattern Recognition (CVPR) Workshops}, June 2019.

\bibitem{LightCNN}
X.~{Wu}, R.~{He}, Z.~{Sun}, and T.~{Tan}.
\newblock A light cnn for deep face representation with noisy labels.
\newblock {\em IEEE Transactions on Information Forensics and Security},
  13(11):2884--2896, 2018.

\bibitem{CasiaSurf}
S.~{Zhang}, A.~{Liu}, J.~{Wan}, Y.~{Liang}, G.~{Guo}, S.~{Escalera}, H.~J.
  {Escalante}, and S.~Z. {Li}.
\newblock Casia-surf: A large-scale multi-modal benchmark for face
  anti-spoofing.
\newblock {\em IEEE Transactions on Biometrics, Behavior, and Identity
  Science}, 2(2):182--193, 2020.

\bibitem{WMCAautoencoder}
Y.~{Zhang}, M.~{Zhao}, L.~{Yan}, T.~{Gao}, and J.~{Chen}.
\newblock Cnn-based anomaly detection for face presentation attack detection
  with multi-channel images.
\newblock In {\em 2020 IEEE International Conference on Visual Communications
  and Image Processing (VCIP)}, pages 189--192, 2020.

\end{thebibliography}
}

\end{document}